\documentclass[journal]{IEEEtran}
\usepackage{graphicx}
\usepackage{tabularx,booktabs}
\newcolumntype{Y}{>{\centering\arraybackslash}X}
\newcolumntype{B}{X}
\newcolumntype{S}{>{\hsize=.5\hsize}X}
\newcolumntype{T}{>{\hsize=.29\hsize}X}
\ifCLASSINFOpdf
\else
\fi
\usepackage{hyperref}
\hypersetup{
    colorlinks=true, 
    linktoc=all,     
    linkcolor=black,
    citecolor=blue 
}

\usepackage{amsmath}
\usepackage{amsfonts} 
\usepackage{array}

\usepackage{cite}

\begin{document}
%
\title{A Survey on Hyperlink Prediction}
%
%
%

\author{Can~Chen,
        Yang-Yu~Liu
\thanks{Manuscript received \today. This work was supported
by the National Institutes of Health (R01AI141529,
R01HD093761, RF1AG067744, UH3OD023268, U19AI095219, and U01HL089856).}
\thanks{C. Chen is with the Channing Division of Network Medicine, Brigham and Women’s Hospital and Harvard Medical School, Boston, MA 02115 USA (spcch@channing.harvard.edu).}
\thanks{Y.-Y. Liu is with the Channing Division of Network Medicine, Brigham and Women’s Hospital and Harvard Medical School, Boston, MA 02115 USA (yyl@channing.harvard.edu).}
}

\maketitle

\begin{abstract}
As a natural extension of link prediction on graphs, hyperlink prediction aims for the inference of missing hyperlinks in hypergraphs, where a hyperlink can connect more than two nodes. Hyperlink prediction has applications in a wide range of systems, from  chemical reaction networks, social communication networks, to protein-protein interaction networks. In this paper, we provide a systematic and comprehensive survey on hyperlink prediction. We propose a new taxonomy to classify existing hyperlink prediction methods into four categories: similarity-based, probability-based, matrix optimization-based, and deep learning-based methods. To compare the performance of methods from different categories, we perform a benchmark study on various hypergraph applications using representative methods from each category. Notably, deep learning-based methods prevail over other methods in hyperlink prediction.
\end{abstract}


\section{Introduction}
Many real-world systems, be they of biological, social, or technological in nature,  can be modeled and analyzed as graphs, where each link (directed or undirected, signed or unsigned, weighted or unweighted) connects two nodes, representing a certain pairwise interaction, association, or physical connection between the two nodes \cite{amaral2000classes, barabasi2003scale, strogatz2001exploring,lindsly20214dnvestigator,lindsly2021functional,sweeney2021network}. For many networked systems (especially biological systems), the discovery and validation of links require significant experimental efforts. Not a big surprise, many real-world networks mapped so far are substantially incomplete. Inferring missing links based on the currently observed network is known as link
prediction, which has tremendous real-world applications in biomedicine \cite{wang2020link,turki2017link}, e-commerce \cite{hasan2011survey,benchettara2010supervised}, social media \cite{wang2015link,tang2015negative}, and criminal intelligence \cite{berlusconi2016link, lim2019hidden}. 

Numerous tools have been developed for predicting or discovering missing or hidden  pairwise interactions (links) in  graphs. Traditional methods include similarity-based methods according to common neighbors \cite{zhou2010solving}, Jaccard index \cite{jaccard1901etude}, and Katz index \cite{katz1953new}. Additionally, advanced deep learning-based methods including deep generative models \cite{wang2020link} and graph convolutional networks (GCN) \cite{zhang2018link} were introduced to tackle the problem. In particular, GCN, which exploits the graph structure to construct neural networks, impressively improves the performance of node/edge classification on graphs compared to traditional neural networks \cite{kipf2016semi, 9046288, bacciu2020gentle}.

\begin{figure}
    \centering
    \includegraphics[scale=1.2]{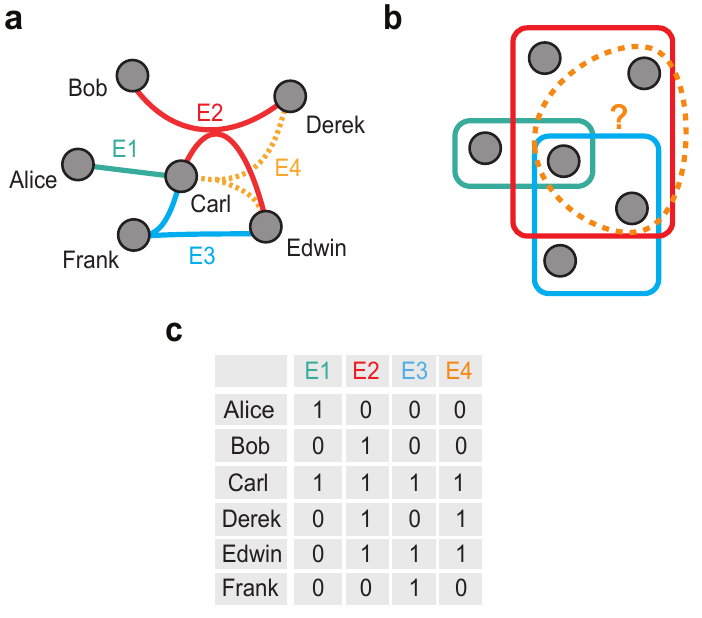}
    \caption{Email communication networks (adapted from \cite{wolf2016advantages}). (a) Schematic representation of an email communication network, where the three solid lines represent the known emails while the orange dash line represents the missing email (we ignore the directionality). (b) Hypergraph representation of the email communication network, where each hyperlink captures the sender and the recipients in the associating email. (c) Incidence matrix of the hypergraph encoded by logical values to indicate the presence or absence of any person in any email.}
    \label{fig:1}
\end{figure}

However, most real-world data representations are multi-dimensional (e.g., co-authorship often involving more than two authors; metabolic reactions often involving more than two metabolites, etc). Using graph models to describe them might result in a loss of higher-order topological features \cite{wolf2016advantages, chen2020tensor, chen2021controllability}. Hypergraphs, a natural generalization of graphs, are superior in modeling the correlation of practical data that could be far more complex than pairwise patterns \cite{gao2020hypergraph}. A hypergraph is composed of hyperlinks (also called hyperedges) which can join any number of nodes. Hypergraphs can represent multi-dimensional relationship naturally and unambiguously \cite{wolf2016advantages, berge1984hypergraphs}. Examples of hypergraphs include email communication networks (Fig. \ref{fig:1} a and b) \cite{wolf2016advantages}, metabolic networks \cite{zhang2018beyond,yadati2020nhp}, co-authorship networks \cite{wolf2016advantages}, actor/actress networks \cite{wolf2016advantages}, and protein-protein interaction networks \cite{klamt2009hypergraphs}. 

The completeness of such large-scale hypergraphs has remained a challenging problem. For instance, even highly curated genome-scale metabolic models have  missing reactions due to our imperfect knowledge of metabolic processes \cite{zimmermann2021gapseq}. Teasing out missing reactions in genome-scale metabolic networks can advance various industrial and biomedicine fields, including metabolic engineering \cite{heinken2021genome, gudmundsson2021recent}, microbial ecology \cite{magnusdottir2017generation}, and drug discovery \cite{robinson2017anticancer}. In addition, predicting potential hyperlinks in hypergraphs is significant for many applications (e.g., predicting collaborations in co-authorship networks) \cite{han2009understanding}. Thus, the development of well-performing methods to recover or predict hyperlinks in hypergraph is imperative.

Hyperlink prediction is an extension of link prediction. The goal of hyperlink prediction is to recover the most likely existent hyperlinks missing from the original hypergraph. Different from link prediction which only deals with pairwise relations, hyperlink prediction is required to find missing hyperlinks with variable cardinality, which significantly increases the difficulty of the problem.  Existing classifiers based on a fixed number of input features become infeasible, and naive generalizations from the link prediction methods often result in a poor performance \cite{zhang2018beyond}. Many efforts have been made in exploring new hyperlink prediction methods. Nevertheless, there is no comprehensive survey on hyperlink prediction methods.

In this paper, we provide a systematic and inclusive literature review of hyperlink prediction methods. We first introduce some preliminary knowledge of hypergraphs and formulate the hyperlink prediction problem in Section \ref{sec:2}. We then create a new taxonomy to classify existing hyperlink prediction methods into four categories, as shown in Section \ref{sec:3}. For each method, we briefly summarize its workflow and discuss its pros and cons. Further, we conduct a benchmark  study of representative methods selected from each category on multiple hypergraph applications, including email communication, school contact, congress bill, drug class, and metabolic networks in Section \ref{sec:4}. The numerical results highlight the effectiveness of the selected methods on various types of hypergraph applications, which will be useful for future hyperlink prediction tasks. Finally, we discuss some potentially interesting directions for future research and conclude in Section \ref{sec:5}. For the ease of reading, we provide a notation table, which includes the most notations used in the paper (Table \ref{tab:1}).

\begin{table}[t]
\centering
\caption{Notations and definitions.}
\begin{tabularx}{8cm}{c *{2}{Y}}
\toprule
Notation & Definition    \\ 
\midrule
$\mathcal{H}$ & Hypergraph \\ 
$\mathcal{H}_{\text{d}}$ & Dual hypergraph \\ 
$\mathcal{V}$ & Node set  \\ 
$\mathcal{E}$ & Hyperlink set \\ 
$\textbf{H}$ & Incidence matrix \\
$d_i$ &  Degree of node $v_i$\\
$c_p$ & Cardinality of hyperlink $e_p$\\
$\textbf{D}$ &  Node degree matrix\\
$\textbf{C}$ & Cardinality degree matrix\\
$\mathcal{N}(v_i)$ & Neighbor set of node $v_i$\\
$\textbf{A}$ & Adjacency matrix\\
$\textbf{S}$ & Pairwise distance matrix\\
$\textbf{L}$ & Normalized Laplacian matrix\\
$\textbf{W}$ & Hyperlink weight matrix \\
$\textsf{L}$ & Normalized Laplacian tensor\\
$\textbf{P}$ & Intersection profile matrix\\
$\textbf{U}$ & Incidence matrix of candidate hyperlinks\\
$\|\|_{\text{F}}$ & Frobenius norm\\
$\sigma$ & Nonlinear activation function\\
$||$ & Vector concatenation\\
Tr & Matrix trace\\
diag & Matrix diagonal\\
\bottomrule
\end{tabularx}
\label{tab:1}
\end{table}

\section{Preliminaries}\label{sec:2}
We briefly review some fundamental concepts of hypergraphs based on the work of \cite{chen2020tensor, chen2021controllability, gao2020hypergraph, berge1984hypergraphs, zhou2006learning, banerjee2017spectra}. A hypergraph is a generalization of graphs in which its hyperlinks (also called hyperedges) can join any number of nodes. Mathematically, an unweighted hypergraph  $\mathcal{H} = \{\mathcal{V}, \mathcal{E}\}$ where $\mathcal{V} = \{v_1,v_2,\dots,v_n\}$ is the node set, $\mathcal{E} = \{e_1,e_2,\dots,e_m\}$ is the hyperlink  set with $e_p\subseteq \mathcal{V}$ for $p=1,2,\dots,m$. Two nodes are called adjacent if they are in the same hyperlink. A hypergraph is called connected if, given any two nodes, there is always a path connecting them through hyperlinks. If all hyperlinks contain the same number of nodes, $\mathcal{H}$ is call a $k$-uniform hypergraph. So a graph is a just 2-uniform hypergraph. Uniform hypergraphs can be naturally and efficiently represented by tensors, i.e., multi-dimensional arrays \cite{chen2020tensor,chen2021controllability,chen2019multilinear, chen2021multilinear, kolda2009tensor}. 

An incidence matrix of a hypergraph, denoted by $\textbf{H}\in\mathbb{R}^{n\times m}$, consists of logical values which indicates the relationship between nodes and hyperlinks (Fig. \ref{fig:1} c). If a node $v_i$ is involved in a hyperlink $e_p$, then the $(i, p)$th entry of \textbf{H}, i.e., $\textbf{H}_{ip}$, has value one. If not, it is equal to zero. The degree of a node is  the number of hyperlinks containing that node, which can be computed as $d_i=\sum_{p}\textbf{H}_{ip}$. We denote the diagonal node degree matrix of a hypergraph by $\textbf{D}\in\mathbb{R}^{n\times n}$. Similarly, the cardinality of a hyperlink is  the number of nodes contained in that hyperlink, which can be computed as $c_p=\sum_{i}\textbf{H}_{ip}$. We denote the diagonal hyperlink cardinality matrix of a hypergraph by $\textbf{C}\in\mathbb{R}^{m\times m}$.

\begin{figure}
    \centering
    \includegraphics[scale=1.1]{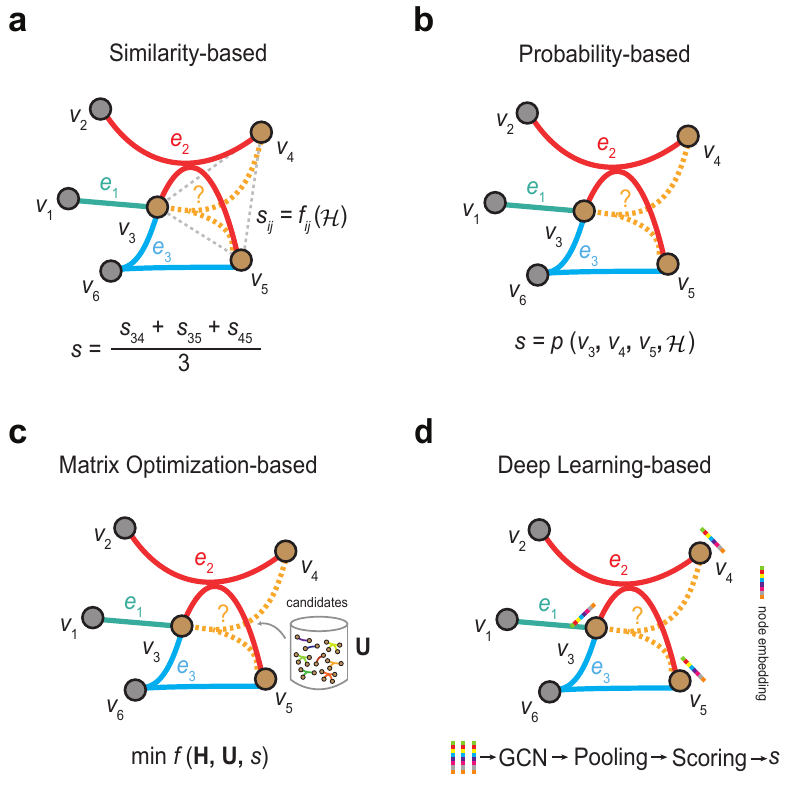}
    \caption{Schematic workflow of the four categories of hyperlink prediction methods.}
    \label{fig:2}
\end{figure}

The goal of hyperlink prediction is to find the most likely existent hyperlinks missing from the observed hyperlink set $\mathcal{E}$. Mathematically, for a given potential hyperlink $e$, most hyperlink prediction methods aim  to learn a function $\Psi$ such that
\begin{equation}
    \Psi(e) = \begin{cases}
    \geq \epsilon \hspace{0.2cm}\text{if $e\in \mathcal{E}$},\\
    < \epsilon \hspace{0.2cm}\text{if $e\notin \mathcal{E}$},
    \end{cases}
\end{equation}
where $\epsilon$ is a threshold to binarize the continuous value of $\Psi$ into a label. In this paper, we systematically review hyperlink prediction methods based on four categories, namely, similarity-based, probability-based, matrix optimization-based, and deep learning-based methods. In each category, we further classify the methods into indirect and direct methods. Indirect methods are those methods initially developed for classification/clustering or other purposes, but can be repurposed for hyperlink prediction. Direct methods are the hypergraph learning methods specifically developed for hyperlink prediction. We list all the hyperlink prediction methods discussed in this paper in Table \ref{tab:2}. A schematic  diagram of each of the four categories is presented in Fig. \ref{fig:2}. 

\begin{table*}[h]
\small
\centering
\caption{A summary of hyperlink prediction methods. Methods marked in bold were used for benchmark evaluations in Section \ref{sec:4}. *: original package for indirect methods (which has not been reformulated for hyperlink prediction); N.A.: not applicable; R.A.: requested from authors.}
\begin{tabularx}{\textwidth}{SSTTTT}
\toprule
\centering  Method & \centering Category & \centering Non-uniform & \centering Direct Method  & \centering\arraybackslash Reference & \centering\arraybackslash Code Avail. \\ 
\midrule
\centering CN  & \centering Similarity & \centering Yes & \centering No  &\centering\arraybackslash \cite{zhou2010solving, zhang2018beyond,kumar2020hpra} & \centering\arraybackslash \href{https://github.com/muhanzhang/HyperLinkPrediction}{GitHub}\\ 
\centering KI & \centering Similarity & \centering Yes & \centering No  &\centering\arraybackslash \cite{katz1953new, zhang2018beyond,kumar2020hpra} & \centering\arraybackslash \href{https://github.com/muhanzhang/HyperLinkPrediction}{GitHub}\\ 
\centering \textbf{HPRA} & \centering Similarity & \centering Yes & \centering Yes  &\centering\arraybackslash \cite{kumar2020hpra} & \centering\arraybackslash \href{https://github.com/darwk/HyperedgePrediction}{GitHub}\\
\midrule
\centering Node2Vec  & \centering Probability & \centering Yes & \centering No & \centering\arraybackslash\cite{grover2016node2vec, yadati2020nhp} & \centering\arraybackslash \href{https://github.com/aditya-grover/node2vec}{GitHub*}\\ 
\centering BS  & \centering Probability & \centering Yes & \centering No  & \centering\arraybackslash\cite{zhang2018beyond,ghahramani2005bayesian} & \centering\arraybackslash \href{https://github.com/muhanzhang/HyperLinkPrediction}{GitHub} \\ 
\centering \textbf{HPLSF} & \centering Probability & \centering Yes & \centering Yes  & \centering\arraybackslash\cite{xu2013hyperlink} & \centering\arraybackslash \href{https://github.com/muhanzhang/HyperLinkPrediction}{GitHub}\\ 
\midrule
\centering FM & \centering Matrix Optimization & \centering Yes &\centering No  & \centering\arraybackslash\cite{zhang2018beyond, rendle2012factorization} & \centering\arraybackslash \href{https://github.com/muhanzhang/HyperLinkPrediction}{GitHub}\\
\centering SHC & \centering Matrix Optimization & \centering Yes & \centering No  & \centering\arraybackslash\cite{zhang2018beyond, zhou2006learning, kumar2020hpra} & \centering\arraybackslash \href{https://github.com/muhanzhang/HyperLinkPrediction}{GitHub} \\
\centering HPTED & \centering Matrix Optimization & \centering No & \centering Yes  & \centering\arraybackslash\cite{maurya2021hyperedge} &\centering\arraybackslash N.A.\\ 
\centering HPLS & \centering Matrix Optimization & \centering Yes & \centering Yes  & \centering\arraybackslash\cite{pan2021predicting} &\centering\arraybackslash N.A.\\ 
\centering MB &\centering Matrix Optimization & \centering Yes & \centering Yes  & \centering\arraybackslash\cite{zhang2016recovering, oyetunde2017boostgapfill} & \centering\arraybackslash \href{https://github.com/Tolutola/BoostGAPFILL}{GitHub}\\ 
\centering CMM & \centering Matrix Optimization & \centering Yes & \centering Yes  &  \centering\arraybackslash\cite{zhang2018beyond} & \centering\arraybackslash \href{https://github.com/muhanzhang/HyperLinkPrediction}{GitHub}\\
\centering \textbf{C3MM} & \centering Matrix Optimization & \centering Yes & \centering Yes  &  \centering\arraybackslash\cite{sharma2021c3mm} & \centering\arraybackslash R.A.\\
\midrule
\centering Node2Vec-SLP & \centering Deep Learning & \centering Yes &\centering No  & \centering\arraybackslash\cite{yadati2020nhp, zhang2019hyper} & \centering\arraybackslash \href{https://pytorch.org/docs/stable/generated/torch.nn.Linear.html}{torch*}\\ 
\centering Node2Vec-GCN & \centering Deep Learning & \centering Yes &\centering No  &\centering\arraybackslash \cite{yadati2020nhp,scarselli2008graph} & \centering\arraybackslash \href{https://pytorch-geometric.readthedocs.io/en/latest/modules/nn.html\#torch\_geometric.nn.conv.GraphConv}{torch\_geometric*}\\ 
\centering Node2Vec-GraghSAGE & \centering Deep Learning & \centering Yes &\centering No  & \centering\arraybackslash\cite{srinivasan2021learning,hamilton2017inductive} & \centering\arraybackslash \href{https://pytorch-geometric.readthedocs.io/en/latest/modules/nn.html\#torch\_geometric.nn.conv.SAGEConv}{torch\_geometric*}\\ 
\centering Node2Vec-RGCN & \centering Deep Learning & \centering Yes &\centering No  & \centering\arraybackslash\cite{srinivasan2021learning, wan2021principled, schlichtkrull2018modeling} & \centering\arraybackslash \href{https://pytorch-geometric.readthedocs.io/en/latest/modules/nn.html\#torch\_geometric.nn.conv.RGCNConv}{torch\_geometric*}\\ 
\centering Node2Vec-HGCN & \centering Deep Learning &\centering Yes & \centering No  & \centering\arraybackslash\cite{yadati2020nhp,feng2019hypergraph, srinivasan2021learning} & \centering\arraybackslash \href{https://github.com/iMoonLab/HGNN}{GitHub*}\\ 
\centering Node2Vec-HyperGCN & \centering Deep Learning &\centering Yes & \centering No  & \centering\arraybackslash\cite{yadati2020nhp,yadati2019hypergcn} & \centering\arraybackslash \href{https://github.com/malllabiisc/HyperGCN}{GitHub*}\\ 
\centering FamilySets & \centering Deep Learning &\centering Yes & \centering Yes  & \centering\arraybackslash\cite{srinivasan2021learning} & \centering\arraybackslash N.A.\\
\centering SNALS & \centering Deep Learning & \centering Yes & \centering Yes  & \centering\arraybackslash\cite{wan2021principled} &\centering\arraybackslash N.A. \\ 
\centering DHNE & \centering Deep Learning & \centering No & \centering Yes  & \centering\arraybackslash\cite{tu2018structural} &  \centering\arraybackslash \href{https://github.com/tadpole/DHNE}{GitHub}\\ 
\centering \textbf{HyperSAGCN} & \centering Deep Learning & \centering Yes & \centering Yes  & \centering\arraybackslash\cite{zhang2019hyper} & \centering\arraybackslash \href{https://github.com/ma-compbio/Hyper-SAGNN}{GitHub}\\ 
\centering \textbf{NHP} & \centering Deep Learning & \centering Yes & \centering Yes  & \centering\arraybackslash\cite{yadati2020nhp} & \centering\arraybackslash R.A.\\ 
\centering \textbf{CHESHIRE} & \centering Deep Learning & \centering Yes & \centering Yes  & \centering\arraybackslash \cite{chen2022teasing} & \centering\arraybackslash \href{https://github.com/canc1993/cheshire-gapfilling}{GitHub}\\ 
\bottomrule
\end{tabularx}
\label{tab:2}
\end{table*}

\section{Methods}\label{sec:3}
In this section, we review existing hyperlink prediction methods, which can be grouped into the following four categories. 

\subsection{Similarity-based Methods}
We discuss three similarity-based methods for hyperlink prediction. The first two (CN, KI) are indirect methods adapted from link prediction, while the last one (HPRA) is a direct method  specifically designed for  hyperlink prediction. Similarity-based methods are always efficient to compute (especially for large hypergraphs compared to other methods), but naive generalizations from similarity-based link prediction methods often result in a poor performance.

\subsubsection{Common Neighbors}
Common neighbors (CN) is a link prediction method that is based on quantifying the overlap or similarity of two nodes in a graph \cite{zhou2010solving}. The similarity index between two nodes $v_i$ and $v_j$ is given by 
\begin{equation}
    \text{CN}_{ij} = |\mathcal{N}(v_i)\cap \mathcal{N}(v_j)|,
\end{equation}
where $\mathcal{N}(v_i)$ denotes the set of neighbors of node $v_i$. CN can be generalized to hyperlinks by calculating the average of the pairwise CN indices between the nodes within each hyperlink \cite{zhang2018beyond, kumar2020hpra,maurya2021hyperedge}, i.e., given a hyperlink $e_p$, the CN index of $e_p$ is given by
\begin{equation}
    \text{CN}_{p} = \frac{2}{c_p(c_p-1)}\sum_{v_i,v_j\in e_p} \text{CN}_{ij}.
\end{equation}

\subsubsection{Katz Index}
Katz index (KI) is a classical similarity measure used for link prediction. It is based on a weighted sum over the collection of all paths connecting nodes $v_i$ and $v_j$, i.e., 
\begin{equation}
    \text{KI}_{ij} = \sum_{l=1}^{\infty} \beta^l (\textbf{A}_\text{g}^l)_{ij}=[(\textbf{I}-\beta \textbf{A}_\text{g})^{-1}-\textbf{I}]_{ij},
\end{equation}
where $\beta$ is a damping factor that gives the shorter paths more weights, $\textbf{A}_\text{g}$ is the adjacency matrix of the target graph, and \textbf{I} is the identity matrix \cite{katz1953new}. KI can be generalized to hyperlinks in the same manner as CN by replacing the graph adjacency matrix $\textbf{A}_\text{g}$ with the hypergraph adjacency matrix \cite{zhang2018beyond, kumar2020hpra}. The adjacency matrix of a hypergraph is often defined as $\textbf{A}=\textbf{H}\textbf{H}^\top-\textbf{D}\in\mathbb{R}^{n\times n}$, which is equivalent to the adjacency matrix of the clique-expanded graph (Fig. \ref{fig:3} b).

\subsubsection{Hyperlink Prediction Using Resource Allocation}
Hyperlink prediction using resource allocation (HPRA) is a recently developed  direct hyperlink prediction method working on the principles of the resource allocation process \cite{kumar2020hpra}. HPRA computes a hypergraph resource allocation (HRA) index between two nodes based on direct connection and common neighbors. Define the direct connection score between node $v_i$ and node $v_j$ as follows:
\begin{equation}
    \text{SD}_{ij} = \sum_{\substack{e_p\ni v_i,v_j\\ i\neq j}} \frac{1}{c_p-1}.
\end{equation}
Then HRA index between the two nodes is given by
\begin{equation}
    \text{HRA}_{ij} = \text{SD}_{ij} + \sum_{v_l\in \mathcal{N}(v_i)\cap \mathcal{N}(v_j)}\frac{\text{SD}_{il}\times\text{SD}_{lj}}{d_l}.
\end{equation}
Similar to CN and KI, the HRA index of a candidate hyperlink can be computed as the average of all the pairwise HRA indices between the nodes within the hyperlink. Moreover, HPRA can predict missing hyperlinks on a hypergraph without knowing any candidate hyperlinks, see details in \cite{kumar2020hpra}.  Different from those naive generalizations, HPRA is able to achieve a good performance on relatively dense hypergraphs while keeping very low computational costs.

\subsection{Probability-based Methods}
We consider three existing probability-based methods for hyperlink prediction. The first two (Node2Vec, BS) are indirect methods, while the last one (HPLSF) is a direct method. Notably, HPLSF is the first machine learning method developed for hyperlink prediction \cite{xu2013hyperlink}. Overall, the probability-based methods often cannot make full use of the structural features of hypergraphs, so the performance is limited.

\subsubsection{Node2Vec}
Node2Vec is a  random walk-based method that learns a mapping of nodes to a low-dimensional space of features that maximizes the likelihood of preserving network neighborhoods of nodes \cite{grover2016node2vec}. Let $f: \mathcal{V} \rightarrow \mathbb{R}^r$ be the mapping function from nodes to feature representations, where $r$ is the dimension of the representations (a hyperparameter chosen by users). Define $\mathcal{N}_S(v_i) \subset \mathcal{V}$ as a network neighborhood of node $v_i$ generated through a neighborhood sampling strategy $S$. Node2Vec  maximizes the log-probability of observing a network neighborhood $\mathcal{N}_S(v_i)$ for node $v_i$ conditioned on its feature representation, i.e., 
\begin{equation}\label{eq:2}
\max_f \sum_{v_i\in\mathcal{V}} \log{\Big(\prod_{v_j\in \mathcal{N}(v_i)}\text{Pr}\big(v_j|f(v_i)\big)\Big)},
\end{equation}
where the conditional probability of every source-neighborhood node pair is defined as
\begin{equation*}
\text{Pr}\big(v_j|f(v_i)\big) = \frac{\exp{(f(v_j)^\top f(v_i))}}{\sum_{v_l\in \mathcal{V}} \exp{\big(f(v_l)^\top f(v_i)\big)}}.
\end{equation*}
The optimization problem (\ref{eq:2}) can be solved by stochastic gradient ascent over the model parameters defining the features $f$. In addition, Node2Vec exploits a flexible biased random walk procedure to explore neighborhoods in a breadth-first sampling as well as depth-first sampling fashion.

\begin{figure*}[t]
    \centering
    \includegraphics[scale=1.3]{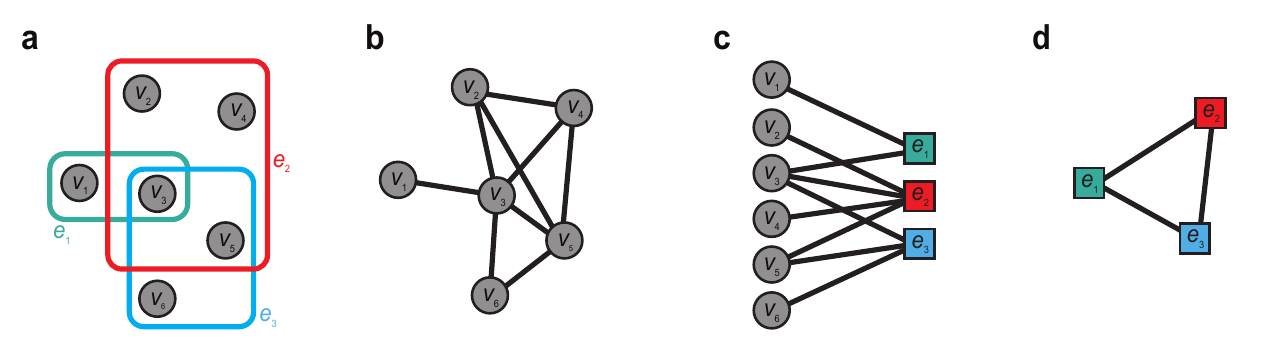}
    \caption{Hypergraph expansion. (a) Original hypergraph. (b) Clique expansion of the hypergraph. (c) Star expansion of the hypergraph (also called the bipartite graph). (d) Line graph of the hypergraph.}
    \label{fig:3}
\end{figure*}

Node2Vec can be applied for hyperlink prediction indirectly \cite{yadati2020nhp}. Given an incomplete hypergraph $\mathcal{H}$, decompose the hypergraph into a graph by clique expansion (Fig. \ref{fig:3} b) and apply Node2Vec on the expanded graph. Suppose that the embedding  of node $v_i$ is $\textbf{x}_i \in \mathbb{R}^r $. Given an unseen hyperlink $e_p$, the hyperlink score can be computed as
\begin{equation}\label{eq:7}
    S_{p} = \text{sigmoid}(\frac{1}{c_p} \sum_{\substack{v_i,v_j\in e_p\\ i\neq j}} \textbf{x}_i^\top\textbf{x}_j),
\end{equation}
which produces a probabilistic metric that measures the average of the correlations between the nodes within $e_p$. The final score $S_p$ therefore can be used to indicate the  existence confidence of $e_p$.  Other  node embedding methods such as DeepWalk \cite{perozzi2014deepwalk} and LINE \cite{tang2015line} can also be used similarly. Node2Vec is a simple and classic method which can be used for hyperlink prediction, but its performance is poor. Decomposing a hypergraph into a graph could lose higher-order structural information. Additionally, Node2Vec is computationally expensive for large dense graphs. Nevertheless, Node2Vec plays an important role in many deep learning-based approaches as the step of embedding initialization.

\subsubsection{Bayesian Set}
Bayesian set (BS) is a probability-based approach for retrieving items from a cluster, given a query consisting of a few items from that cluster, as a Bayesian inference problem \cite{ghahramani2005bayesian}. The method utilizes a model-based concept of a cluster and ranks items using a score which evaluates the marginal probability that each item belongs to a cluster containing the query items. Let $D$ be a data set of items and $D_c$ be a query set with $D_c\subset D$. Having observed $D_c$, the score for an item $x\in D$ belonging $D_c$ is given by
\begin{equation}
    S(x) = \frac{p(x|D_c)}{p(x)} = \frac{p(x, D_c)}{p(x)p(D_c)}.
\end{equation}
The numerator represents the probability that $x$ and $D_c$ are generated by the same model with the same parameters, while the denominator represents  the probability that $x$ and $D_c$ came from models with different parameters. BS has been used in the setting of hyperlink prediction, where $D_c$ and $D$ can be viewed as known hyperlink set and all candidate hyperlink set, respectively \cite{zhang2018beyond}. However, since BS is an information retrieval method that only retrieves similar items, it does not perform well on hyperlink prediction tasks. 

\IEEEpeerreviewmaketitle

\subsubsection{Hyperlink Prediction Using Latent Social Features}
Hyperlink prediction using latent social features (HPLSF) is the first hyperlink prediction method in the hypergraph learning community, which exploits the homophily property of social networks and introduces a latent feature learning scheme \cite{xu2013hyperlink}. HPLSF can be categorized as either probability-based or matrix optimization-based. We here treat it as a probability-based method due to the novelty of utilizing entropy in computing hyperlink embeddings. Given an incomplete hypergraph $\mathcal{H}$ with $n$ nodes, denote $\textbf{S}\in\mathbb{R}^{n\times n}$ as the distance matrix of $\mathcal{H}$ (i.e., $\textbf{S}_{ij}$ is the length of the shortest path from node $v_i$ to node $v_j$). HPLSF first finds the latent features through multi-dimensional scaling (MDS), i.e., 
\begin{equation}
    \min_{\textbf{Z}} \|\textbf{S} - \textbf{Z}\textbf{Z}^\top\|_{\text{F}},
\end{equation}
where $\textbf{Z}\in\mathbb{R}^{n\times k}$ is the latent feature matrix. Subsequently, HPLSF computes an entropy-based embedding for each hyperlink based on the latent node features (similar to an entropy-based pooling function). Given a hyperlink $e_p$, the entropy-based embedding of $e_p$ is then given by
\begin{align}
    \textbf{y}_p = \begin{bmatrix} \gamma(\textbf{Z}^{(p)}_{:1}) & \gamma(\textbf{Z}^{(p)}_{:2}) & \dots & \gamma(\textbf{Z}^{(p)}_{:k})\end{bmatrix}\in\mathbb{R}^k,
\end{align}
where $\textbf{Z}^{(p)}\in\mathbb{R}^{c_p\times k}$ is the latent feature matrix for the nodes contained in $e_p$ ($\textbf{Z}^{(p)}_{:j}$ represents the $j$th column of $\textbf{Z}^{(p)}$), and $\gamma$ computes the Shannon entropy of a vector, i.e., for $j=1,2,\dots,k$,
\begin{equation}
\gamma(\textbf{Z}^{(p)}_{:j})= -\sum_{i=1}^{c_p} \frac{\textbf{Z}^{(p)}_{ij}}{\sum_{l=1}^{c_p}\textbf{Z}^{(p)}_{lj}}\ln{\frac{\textbf{Z}^{(p)}_{ij}}{\sum_{l=1}^{c_p}\textbf{Z}^{(p)}_{lj}}}. 
\end{equation}
If the observed node feature matrix is also provided, HPLSF will repeat the same entropy calculation to obtain another entropy-based embedding of $e_p$. Finally, HPLSF trains a structural support vector machine model with combined observed and latent hyperlink embeddings to perform hyperlink prediction. 

In the work of \cite{zhang2018beyond}, the authors modified HPLSF by training a logistic regression model on the latent entropy-based hyperlink embeddings in order to output prediction scores. HPLSF only considers the pairwise distances between nodes when generating latent node features without including any higher-order topological attributes. Thus, more advanced techniques which can fully exploit hypergraph structure are required for improving the performance of hyperlink prediction.

\subsection{Matrix Optimization-based Methods}
We investigate several matrix optimization-based hyperlink prediction methods. The first two (FM, SHC) are indirect methods, while the remaining (HPTE, HPLS, MB, CMM, C3MM) are direct methods. The essence of these methods is to exploit the incidence, adjacency, or Laplacian matrices (or Laplacian tensors) of hypergraphs to formulate matrix optimization problems for hyperlink prediction.

\subsubsection{Factorization Machine}
Factorization machine (FM) integrates the generality of feature engineering with the superiority of factorization models in estimating interactions between variables of large domain \cite{rendle2012factorization}. Suppose that the input feature variable is $\textbf{x}\in\mathbb{R}^r$.  The FM model of order two is defined as
\begin{equation}\label{eq:12}
    y(\textbf{x}) = w_0 + \sum_{i=1}^r w_i x_i + \sum_{i=1}^r\sum_{j=i+1}^r x_ix_j\sum_{f=1}^ks_{if}s_{jf},
\end{equation}
where $\Theta=\{w_0,w_1,\dots,w_r, s_{11},\dots,s_{rk}\}$ is the set of model parameters, and $x_i$ denotes the $i$th entry of $\textbf{x}$. The first two terms of the FM model contain the unary interactions of each input variable $x_i$ with the target, which is equivalent to the linear regression model. The last term with the two nested sums contains all pairwise interactions of the input variables.
Optimality of the model parameters is defined with a loss function (e.g., least-square or cross-entropy loss) where the goal is to minimize the sum of losses over the observed data. Given an incomplete hypergraph $\mathcal{H}$, FM  treats hyperlink prediction as a simple classification problem by fitting (\ref{eq:12}) with the incidence matrix of $\mathcal{H}$  \cite{zhang2018beyond}. Although the incidence matrix of a hypergraph encoded with higher-order relations, the performance of FM is poor due to its simple learning scheme. 

\subsubsection{Spectral Hypergraph Clustering}
Spectral hypergraph clustering (SHC) is the first semi-supervised hypergraph learning method developed for label prediction on hypergraphs \cite{zhou2006learning}. SHC aims to learn a partition in which the connections among the nodes within the same part are dense while the connections between two parts are sparse. SHC generalizes the powerful methodology of spectral clustering, which originally operates on graphs to hypergraphs. Given a hypergraph $\mathcal{H}$ with $n$ nodes, let $\textbf{f}\in\mathbb{R}^n$ be the classification function and $\textbf{y}\in\mathbb{R}^n$ be the label vector consisting of values of 0, 0.5, and 1 where 0.5 indicates those unlabeled nodes. The whole SHC learning model can be defined as
\begin{equation}\label{eq:8}
    \min_{\textbf{f}} \text{ }  \|\textbf{f}-\textbf{y}\|_{\text{F}}^2 + \mu\textbf{f}^\top\textbf{L}\textbf{f},
\end{equation}
where $\mu>0$ is the regularization parameter, and $\textbf{L}$ is the normalized Laplacian matrix of the hypergraph defined by
\begin{equation}\label{eq:9}
    \textbf{L} =  \textbf{I} - \textbf{D}^{-\frac{1}{2}}\textbf{H}\textbf{W}\textbf{C}^{-1}\textbf{H}^\top\textbf{D}^{-\frac{1}{2}}\in\mathbb{R}^{n\times n}.
\end{equation}
Here $\textbf{I}\in\mathbb{R}^{n\times n}$ is the identity matrix, and $\textbf{W}\in\mathbb{R}^{m\times m}$ is a diagonal matrix of hyperlink weights ($\textbf{W}=\textbf{I}$ if $\mathcal{H}$ is unweighted). The normalized Laplacian matrix $\textbf{L}$ in fact can be viewed as the normalized Laplacian matrix of the clique-expanded graph with edge weights scaled by the associated hyperlink cardinality. The hypergraph regularizer $\textbf{f}^\top\textbf{L}\textbf{f}$ implies that the state of a node is more affected by its neighborhoods with closer and stronger connections than those remote nodes. The optimization problem (\ref{eq:8}) can be solved in a closed-form, i.e., 
\begin{equation}
    \textbf{f} = (\textbf{I}-\mu \textbf{L})^{-1}\textbf{y}.
\end{equation}

SHC can be used to predict missing hyperlinks by transforming the target hypergraph to its dual hypergraph \cite{zhang2018beyond, kumar2020hpra, yadati2020nhp}. Given an incomplete hypergraph $\mathcal{H}$ with $n$ nodes and $m$ hyperlinks, its dual hypergrpah, denoted by $\mathcal{H}_\text{d}$, can be simply obtained by switching the node and hyperlink relations. The incidence matrix of $\mathcal{H}_\text{d}$ is then given by $\textbf{H}_\text{d} = \textbf{H}^\top\in\mathbb{R}^{m\times n}$. Additionally, more advanced hypergraph spectral clustering methods such as dynamic hypergraph structure learning  \cite{zhang2018dynamic}, tensor-based dynamic hypergraph structure learning  \cite{gao2020hypergraph}, hypergraph label propagation network  \cite{zhang2020hypergraph}, and nonlinear diffusion method  \cite{tudisco2021nonlinear} can be applied for hyperlink prediction with a similar manner. SHC has achieved a reasonable performance in hyperlink prediction due to utilization of hypergraph structure. However, SHC is a simple method initially designed for node classification/clustering by leveraging hyperlink relations. Converting a hypergraph to its dual could still lead to a loss of structural features.

\subsubsection{Hyperlink Prediction Using Tensor Eigenvalue Decomposition}
Hyperlink prediction using tensor eigenvalue decomposition (HPTED) utilizes the Fiedler eigenvector, computed using tensor eigenvalue decomposition of the hypergraph Laplacian tensor, to conduct hyperlink prediction \cite{maurya2021hyperedge}. A tensor is a multi-dimensional array generalized from vectors and matrices. The order of a tensor is the number of its dimensions. A $k$-uniform hypergraph $\mathcal{H}$ with $n$ nodes can be naturally represented by a $k$th-order supersymmetric tensor (invariant under permutation of the indices) of size $n$ in each dimension, i.e., $\mathbb{R}^{n\times n\times \stackrel{k}{\dots}\times n}$  \cite{chen2020tensor,chen2021controllability}. The normalized Laplacian tensor of $\mathcal{H}$ is  defined by
\begin{equation}
    \textsf{L}_{i_1i_2\dots i_k} = 
    \begin{cases}
    \frac{-\textbf{W}_{pp}}{(k-1)!\prod_{l=1}^k\sqrt[k]{d_{i_l}}}  \text{ if } i_1,i_2,\dots,i_k\in e_p\\
    1 \hspace{0.2cm}\text{ if } i_1=i_2=\dots=i_k\\
   0 \hspace{0.2cm}\text{ otherwise},
    \end{cases}
\end{equation}
where $\textbf{W}$ is a diagonal matrix of hyperlink weights (if $\mathcal{H}$ is unweighted, $\textbf{W}_{pp}=1$) \cite{banerjee2017spectra}. The Fiedler eigenvector is the eigenvector corresponding to the minimum positive eigenvalue (Fielder value) from the tensor eigenvalue decomposition of $\textsf{L}$, which can be solved by matrix/tensor optimization. Details of the optimization  for solving tensor eigenvalues/eigenvectors can be found in \cite{maurya2021hyperedge, qi2018tensor, chen2017fiedler}. After obtaining the Fiedler eigenvector, HPTED computes a construction cost of a potential hyperlink $e_p$, i.e., 
\begin{equation}
    l_{p}(\textbf{x}) = \textbf{W}_{pp}(\sum_{i_l\in e_p}x_{i_l}^k - k \prod_{i_l\in e_p}x_{i_l}),
\end{equation}
where $x_{i_l}$ is the $i_l$th entry of the Fiedler eigenvector $\textbf{x}\in\mathbb{R}^n$. It has proved that the construction cost $l_p$ represents the contribution of $e_p$ to the Fielder value of the hypergraph Laplacian, which quantifies the connectivity of the hypergraph \cite{maurya2020hypergraph}. Therefore, a smaller construction cost indicates higher existence confidence of $e_p$. In other words, HPTED can be viewed as the inclusion of new hyperlinks such that there is minimal perturbation to the connectivity of the hypergraph.

Although HPTED successfully keeps the higher-order structural features of a hypergraph using tensor theory, the method is not applicable to non-uniform hypergraphs, and most real-world hypergraphs are non-uniform. More importantly, computing the tensor eigenvalues and eigenvectors of a tensor is NP-hard \cite{hillar2013most}. Current computation schemes  such as \cite{chen2016computing} would become intractable when dealing with large tensors.

\subsubsection{Hyperlink Prediction via Loop Structure}
Hyperlink prediction via loop structure (HPLS) exploits the loop features of a hypergraph to perform hyperlink prediction \cite{pan2021predicting}.  HPLS can be categorized as either probability-based or matrix optimization-based. We here treat it as a matrix optimization-based approach due to the novelty of using adjacency and intersection profile matrices in quantifying hypergraph loop features. There are two types of loops used in HPLS -- node-based loops (walks that start and end at the same node) and hyperlink-based loops (walks that start and end at the same hyperlink). Given an incomplete hypergraph $\mathcal{H}$ with $n$ nodes and $m$ hyperlinks, denote $\textbf{A}=\textbf{H}\textbf{H}^\top-\textbf{D}\in\mathbb{R}^{n\times n}$ and $\textbf{P}=\textbf{H}^\top\textbf{H}-\textbf{C}\in\mathbb{R}^{m\times m}$ as the adjacency matrix and the intersection profile matrix of $\mathcal{H}$, respectively. Then the weighted sum over loops with different lengths is defined as
\begin{equation}
    SL(\mathcal{H}) = \sum_{\tau=2}^{\tau_c} \alpha_{\tau} \log{\Big(\text{Tr}(\textbf{A}^{\tau})\Big)} + \sum_{\tau=2}^{\tau_c} \beta_{\tau} \log{\Big(\text{Tr}(\textbf{P}^{\tau})\Big)},
\end{equation}
where $\alpha_\tau, \beta_\tau$ are the weight parameters, $\tau_c$ is the cuttoff of the loop length, and $\text{Tr}(\textbf{A}^{\tau})$ and $\text{Tr}(\textbf{P}^{\tau})$ are the total numbers of node-based loops and hyperlink-based loops of length $\tau$, respectively (Tr denotes the matrix trace operation). Given a potential hyperlink $e_p$, define two hypergraphs $\mathcal{H}_{p+}=\{\mathcal{V},\mathcal{E}\cup \{e_p\}\}$ and $\mathcal{H}_{p-}=\{\mathcal{V}, \mathcal{E}\backslash \{e\}\}$. Let $S_p$ be the probability of existence for $e_p$, and its log-odds is assumed by
\begin{equation}
    \log{\frac{S_p}{1-S_p}} = c + \frac{1}{c_p^{\gamma}}\Big (SL(\mathcal{H}_{p+})-SL(\mathcal{H}_{p-})\Big),
\end{equation}
where $c$ and $\gamma$ are parameters to be determined. Finally, HPLS maximizes the following likelihood for obtaining the optimal parameter set $\{\alpha_\tau,\beta_\tau,\gamma\}$:
\begin{equation}
    \max_{\alpha_\tau,\beta_\tau,\gamma}{\prod_{e_p\in \mathcal{E}\cup \mathcal{F}}} S_p^{\mathbb{I}(e_p\in \mathcal{E})}(1-S_p)^{1-\mathbb{I}(e_p\in \mathcal{E})},
\end{equation}
where $\mathcal{F}$ is a negative hyperlink set, and $\mathbb{I}$ is an indicator function. HPLS has achieved a better performance compared to previous methods such as Katz, BS, and SHC. However, it is not computationally efficient for large hypergraphs. The matrix multiplication of \textbf{A} and \textbf{P} for $\tau_c-1$ times is extremely time-consuming.

\subsubsection{Matrix Boost}
The remaining three methods in this category are a series of matrix optimization-based hyperlink prediction methods. Matrix boost (MB) conducts inference jointly in the incidence and adjacency space by performing an iterative completion-matching optimization \cite{zhang2016recovering}. MB has been successfully applied to predict missing reactions in genome-scale metabolic networks \cite{oyetunde2017boostgapfill}. Given an incomplete hypergraph $\mathcal{H}$ with $n$ nodes, denote $\textbf{A}=\textbf{H}\textbf{H}^\top\in\mathbb{R}^{n\times n}$ as the adjacency matrix of $\mathcal{H}$ (defined sightly different from that in HPLS by including self-loops). Suppose that the complete adjacency matrix is given by $\textbf{A}+\Delta\textbf{A}$, and it can be decomposed by
\begin{equation}
    \textbf{A}+\Delta\textbf{A} = \textbf{A} + [\Delta\textbf{A}]_{\textbf{A}} + [\Delta\textbf{A}]_{\bar{\textbf{A}}},
\end{equation}
where  $[\textbf{X}]_{\textbf{A}}$ denotes the operation that only keeps the entries of $\textbf{X}$ at $\textbf{A}$’s nonempty entries and mask all else, and $[\textbf{X}]_{\bar{\textbf{A}}}$ is conversely defined as keeping $\textbf{X}$ only at $\textbf{A}$’s empty entries. Define $\textbf{A} + [\Delta\textbf{A}]_{\textbf{A}}=\textbf{A}^{+}$ and $[\Delta\textbf{A}]_{\bar{\textbf{A}}}=\Delta\textbf{A}^{-}$. MB first aims to approximate the empty entries of $\textbf{A}^{+}$, denoted by $\Delta \hat{\textbf{A}}$, with known $\textbf{A}^{+}$ (which can also be  approximated iteratively). The optimization problem is as follows:
\begin{equation}
        \min_{\boldsymbol{\Theta}} \sum_{i<j} \|\textbf{A}^{+}_{ij} - y_{ij}\|_{\text{F}}^2 + \gamma \mathcal{R}(\boldsymbol{\Theta}),
\end{equation}
where $\boldsymbol{\Theta}=\{w_0, w_i, w_j, s_{if}, s_{jf}\}$ is the set of parameters, $y_{ij}=w_0+w_i+w_j+\sum_{f=1}^k s_{if}s_{jf}$, and $\mathcal{R}$ is a regularizer. After training, $\Delta\hat{\textbf{A}}$ can be obtained by
\begin{equation}
\Delta\hat{\textbf{A}}_{ij} = 
    \begin{cases}
    w_0+w_i+w_j + \sum_{f}s_{if}s_{jf} \text{ if } \textbf{A}^{+}=0,\\
    0 \hspace{0.2cm}\text{ if } \textbf{A}^{+}\neq 0.
    \end{cases}
\end{equation}
Let $\textbf{U}\in\mathbb{R}^{n\times \tilde{m}}$ be the incidence matrix of the candidate hyperlinks of $\mathcal{H}$ and $\boldsymbol{\Lambda}\in\mathbb{R}^{\tilde{m}\times \tilde{m}}$ be a diagonal indicator matrix of the candidate hyperlinks. In the matching step, MB solves the optimization problem as follows:
\begin{equation}\label{eq:20}
\begin{split}
        \min_{\boldsymbol{\Lambda}} \text{ } & \|[\textbf{U}\boldsymbol{\Lambda}\textbf{U}^\top]_{\bar{\textbf{A}}} - \Delta \hat{\textbf{A}}\|_{\text{F}}^2\\
        \text{subject to } & \boldsymbol{\Lambda}_{pp} = \{0,1\} \text{ for } p = 1,2,\dots,\tilde{m}.
\end{split}
\end{equation}
The optimization problem (\ref{eq:20}) can be relaxed by making the integer $\boldsymbol{\Lambda}_{pp}$ continuous within $[0, 1]$, which can be solved by subgradient methods \cite{boyd2003subgradient}. The continuous scores $\boldsymbol{\Lambda}_{pp}$ can be viewed as soft indicators of the candidate hyperlinks. 

MB leverages the powerful matrix factorization technique to perform inference in the adjacency space in recovering missing hyperlinks. Yet, it has limited scalibility since the candidate hyperlink set must be present during training. If the candidate hyperlink set becomes extremely large (e.g., the entire metabolic reaction universe), the matrix optimization will be difficult (or even impossible) to solve. Moreover, MB cannot handle unseen hyperlinks at test time, which limits the applications of the method.

\subsubsection{Coordinated Matrix Minimization}
Coordinated matrix minimization (CMM) is an improved version of MB, which introduces a latent factor matrix to significantly simplify the method \cite{zhang2018beyond}. CMM alternatively performs non-negative matrix factorization and least square matching in the adjacency space, in order to infer a subset of candidate hyperlinks that are most suitable to fill the target hypergraph. Similarly to MB, denote  $\textbf{A}=\textbf{H}\textbf{H}^\top\in\mathbb{R}^{n\times n}$ and $\textbf{U}\in\mathbb{R}^{n\times \tilde{m}}$ as the adjacency matrix  of $\mathcal{H}$ and the incidence matrix of the candidate hyperlinks, respectively. Let a non-negative matrix $\textbf{Q}\in\mathbb{R}^{n\times k}$ be the latent factor matrix ($k\ll n$), and assume that the complete adjacency matrix of the hypergraph is factoried by 
\begin{equation}
    \textbf{A} + \textbf{U}\boldsymbol{\Lambda}\textbf{U}^\top \approx \textbf{Q}\textbf{Q}^\top,
\end{equation}
where $\boldsymbol{\Lambda}\in\mathbb{R}^{\tilde{m}\times \tilde{m}}$ is a diagonal indicator matrix of candidate hyperlinks. To find the missing hyperlinks, CMM solves the following optimization problem by using the expectation–maximization algorithm:
\begin{equation}\label{eq:cmm}
\begin{split}
    \min_{\boldsymbol{\Lambda}, \textbf{Q}\geq 0} \text{ } & \|\textbf{A}+\textbf{U}\boldsymbol{\Lambda}\textbf{U}^\top-\textbf{Q}\textbf{Q}^\top\|^2_{\text{F}}\\
    \text{subject to } &\boldsymbol{\Lambda}_{pp} = \{0,1\} \text{ for } p = 1,2,\dots,\tilde{m}.
\end{split}
\end{equation}
After relaxing the constraint of $\boldsymbol{\Lambda}_{pp}$ to be continuous within $[0,1]$, the linear least square problem  can be solved very efficiently using off-the-shelf optimization tools, e.g., IBM CPLEX \cite{manual1987ibm}. Although CMM is simpler than MB with a better performance, it still suffers from the issue of scalability and cannot handle unseen hyperlinks. 

\subsubsection{Clique Closure-based Coordinated Matrix Minimization}
Clique closure-based coordinated matrix minimization (C3MM) is an improved version of CMM \cite{sharma2021c3mm}. C3MM introduces a clique closure hypothesis (i.e., hyperlinks are more likely to be formed from near-cliques rather than from non-cliques) into the objective function of CMM, which significantly hunts down more hyperlinks which are missed by CMM. C3MM first approximates the latent factor matrix $\textbf{Q}\in\mathbb{R}^{n\times k}$ ($k\ll n$). Given a diagonal indicator matrix $\boldsymbol{\Lambda}_{\textbf{U}}\in\mathbb{R}^{\tilde{m}\times \tilde{m}}$ (which can be initialized randomly), C3MM computes
\begin{equation}
    \min_{\textbf{W}\geq 0} \|\textbf{A} + \textbf{A}_{\text{CN}} + \textbf{U}\boldsymbol{\Lambda}_{\textbf{U}}\textbf{U}^\top - \textbf{Q}\textbf{Q}^\top\|_{\text{F}}^2,
\end{equation}
where $\textbf{A}_{\text{CN}} = \textbf{A}^2-\text{diag}(\textbf{A}^2)$ captures the common neighbor information of the projected graph (``diag" denotes the diagonal operation that keeps the diagonal of a matrix with zero elsewhere). Define $\Delta \textbf{A} = \textbf{Q}\textbf{Q}^\top-\textbf{A}$. To find missing hyperlinks, C3MM solves the second optimization problem as follow:
\begin{equation}
\begin{split}
    \min_{\boldsymbol{\Lambda}_{\textbf{U}}, \boldsymbol{\Lambda}_{\textbf{H}}} & \|\textbf{A} - \textbf{H} \boldsymbol{\Lambda}_{\textbf{H}}\textbf{H}^\top - \textbf{U}\boldsymbol{\Lambda}_{\textbf{U}}\textbf{U}^\top\|_{\text{F}}^2 \\ &+ \|\Delta\textbf{A}-\textbf{U}\boldsymbol{\Lambda}_{\textbf{U}}\textbf{U}^\top\|_{\text{F}}^2 + \|\boldsymbol{\Lambda}_{\textbf{H}}\|_1\\
    \text{subject to }&(\boldsymbol{\Lambda}_{\textbf{U}})_{pp} = \{0,1\} \text{ for } p = 1,2,\dots,\tilde{m} \\ & (\boldsymbol{\Lambda}_{\textbf{H}})_{pp} = \{0,1\} \text{ for } p = 1,2,\dots,m.
\end{split}
\end{equation}
The method solves the two optimization problems alternatively for a certain number of iterations. C3MM has proved to perform well on temporal hyperlink prediction tasks, compared to CMM. However, C3MM has the same issues with MB and CMM (i.e., scalability and inability of handling unseen hyperlinks). Therefore, more sophisticated deep learning techniques are needed in order to fix these issues.  

\subsection{Deep Learning-based Methods}
We explore the existing literature regarding deep learning-based methods for hyperlink prediction. The first six (Node2Vec-SLP, Node2Vec-GCN, Node2Vec-GraphSAGE, Node2Vec-RGCN, Node2Vec-HGCN, Node2Vec-HyperGCN) are indirect methods, while the last six (FamilySet, SNALS, DHNE, Hyper-SAGCN, NHP, CHESHIRE) are direct methods. In particular,  utilization of graph/hypergraph-based neural networks significantly improves the performance of hyperlink prediction. 

\subsubsection{Node2Vec with Single-layer Perceptron}
Node2Vec with single-layer perceptron (Node2Vec-SLP) is an improved version of Node2Vec for hyperlink prediction, which employs a one-layer neural network to compute hyperlink scores \cite{yadati2020nhp, zhang2019hyper}. Given an incomplete hypergraph $\mathcal{H}$, decompose the hypergraph into a graph by clique expansion and apply Node2Vec on the expanded graph. Suppose that the embedding of node $v_i$ is $\textbf{x}_i$. Thus, the embedding of hyperlink $e_p$, denoted by $\textbf{y}_p$, can be obtained by aggregating all the embeddings of the nodes within the hyperlink through a pooling function. Many pooling functions can be used such as maximum pooling, minimum pooling, and mean pooling. The embedding of $e_p$ is further fed into a one-layer neural network to produce a probabilistic score, i.e.,
\begin{equation}\label{eq:25}
    S_{p} = \text{sigmoid}(\textbf{W}_{\text{score}}\textbf{y}_p+\textbf{b}_{\text{score}}),
\end{equation}
where $\textbf{W}_{\text{score}}$ and $\textbf{b}_{\text{score}}$ are learnable parameters in the neural network. The final score $S_p$ can be used ot indicate the  existence confidence of $e_p$. Node2Vec-SLP sightly improves the performance of Node2Vec in terms of hyperlink prediction, but it does not fundamentally solve the issues carried from Node2Vec (i.e., losing higher-order structural information). 

\subsubsection{Node2Vec with Graph Convolutional Network}
Node2Vec with graph convolutional network (Node2Vec-GCN) is an extension of Node2Vec-SLP by introducing an embedding refinement step before hyperlink pooling \cite{yadati2020nhp}. In particular, the embedding refinement is defined by a GCN constructed based on the clique expansion of $\mathcal{H}$. Suppose that the embedding of node $v_i$ is $\textbf{x}_i$ (generated by Node2Vec). Then the refined embedding of $v_i$ through a GCN layer is given by
\begin{equation}\label{eq:26}
    \hat{\textbf{x}}_i = \sigma\Big(\textbf{W}_{\text{conv1}} \textbf{x}_i + \sum_{v_j\in \mathcal{N}(v_i)}\textbf{W}_{\text{conv2}}\textbf{x}_j\Big),
\end{equation}
where $\mathcal{N}(v_i)$ denotes the neighbor set of node $v_i$ in the expanded graph, $\sigma$ is a nonlinear activation function, and $\textbf{W}_{\text{conv1}}$ and $\textbf{W}_{\text{conv2}}$ are  learnable parameters in the GCN \cite{scarselli2008graph}. The remaining steps of Nod2Vec-GCN follow Node2Vec-SLP.

\subsubsection{Node2Vec with GraphSAGE}
Node2Vec with GraphSAGE (Node2Vec-GraphSAGE) is an alternative to Node2Vec-GCN by replacing the GCN layer with a GraphSAGE layer \cite{srinivasan2021learning}. The refined embedding of $v_i$ through a GraphSAGE layer is given by
\begin{equation}
     \hat{\textbf{x}}_i = \sigma\Big(\textbf{W}_{\text{conv}}(\textbf{x}_i || \frac{1}{|\mathcal{N}(v_i)|}\sum_{v_j\in \mathcal{N}(v_i)}\textbf{x}_j)\Big), 
\end{equation}
where $``||"$ denotes the vector concatenation operation, and $\textbf{W}_{\text{conv}}$ is a learnable parameter in the GraphSAGE. Other graph-based neural networks such as \cite{defferrard2016convolutional, bresson2017residual, velivckovic2017graph, shi2020masked} can also be used for embedding refinement. However, both Node2Vec-GCN and Node2Vec-GraphSAGE use the clique-expanded graph structure to refine node embeddings, which has the issue of losing higher-order structural information. 

\subsubsection{Node2Vec with Relational Graph Convolutional Network}
Instead of refining node embeddings on the clique-expanded graph, Node2Vec with relational graph convolutional network (Node2Vec-RGCN) exploits star expansion (Fig. \ref{fig:3} c) with a RGCN for updating node embeddings \cite{srinivasan2021learning, wan2021principled}. RGCN was developed specifically to deal with knowledge graphs where edges have different types \cite{schlichtkrull2018modeling}. Suppose that the embedding of node $v_i$ is $\textbf{x}_i$ (generated by Node2Vec). After obtaining the bipartite graph, Node2Vec-RGCN updates the node embeddings as follows:
\begin{equation}
    \begin{split}
        \textbf{y}_p &= \sigma(\sum_{v_i\in e_p}\textbf{W}_{\text{conv1}} \textbf{x}_i),\\
        \hat{\textbf{x}}_i & = \sigma(\sum_{e_p \ni v_i} \textbf{W}_{\text{conv2}}\textbf{y}_p),
    \end{split}
\end{equation}
where $\textbf{W}_{\text{conv1}}$ and $\textbf{W}_{\text{conv2}}$ are  learnable parameters in the RGCN. The remaining steps of Node2Vec-RGCN follow Node2Vec-SLP. Although the star expansion of a hypergraph somehow preserves higher-order structural features, Node2Vec-RGCN fails to capture node-to-node and hyperlink-to-hyperlink interactions. 

\subsubsection{Node2Vec with Hypergraph Convolutional Network}
Hypergraph convolutional network (HGCN), a convolutional neural network built directly on hypergraphs, is able to learn the hidden layer representation considering the high-order data structure  \cite{feng2019hypergraph}. Experiments have shown that HGCN outperforms graph-based neural networks on hypergraph data. HGCN can be applied for hyperlink prediction indirectly, similar to those graph-based neural networks \cite{yadati2020nhp, srinivasan2021learning, wan2021principled}. Suppose that the embeddings of all the nodes of $\mathcal{H}$ are represented by $\textbf{X}$ (generated by Node2Vec). Then the embedding refinement defined by a HGCN layer is given by 
\begin{equation}
    \hat{\textbf{X}} =\sigma(\textbf{L}\textbf{X} \textbf{W}_{\text{conv}}), 
\end{equation}
where $\textbf{L}$ is the normalized Laplacian matrix of $\mathcal{H}$ defined in (\ref{eq:9}), and $\textbf{W}_{\text{conv}}$ is a learnable parameter in the HGCN. The remaining steps of Node2Vec-HGCN follow Node2Vec-SLP. As mentioned, \textbf{L} can be viewed as the normalized Laplacian matrix of the clique-expanded graph with edge weights scaled by the associated hyperlink cardinality. Thus, the improvement of Node2Vec-HGCN is limited, compared to the previous deep learning-based methods. 

\subsubsection{Node2Vec with HyperGCN}
HyperGCN is an another newly-developed convolutional network on hypergraphs, which has achieved a better performance compared to HGCN on node classification \cite{yadati2019hypergcn}. The key of HyperGCN is to construct a projected graph $\mathcal{G}$ while keeping the higher-order topological features from $\mathcal{H}$. Suppose that the embedding  of node $v_i$ is $\textbf{x}_i$ (generated by Node2Vec). For each hyperlink $e_p$, define an edge between node $v_i\in e_p$ and node $v_j\in e_p$ such that
\begin{equation}\label{eq:29}
    (v_i,v_j)=\text{argmax}_{\substack{v_i,v_j\in e_p\\ i\neq j}}\|\textbf{Q}(\textbf{x}_i-\textbf{x}_j)\|_{\text{F}},
\end{equation}
where $\textbf{Q}$ is a learnable weight matrix. Subsequently, connect the selected two nodes $v_i$ and $v_j$ with the remaining nodes in $e_p$ and set the weight of each edge to $\frac{1}{2c_p-3}$. After obtaining  $\mathcal{G}$, Node2Vec-HyperGCN utilizes the GCN defined in (\ref{eq:26}) for node embedding refinement \cite{yadati2020nhp}. The weight matrix $\textbf{Q}$ defined in (\ref{eq:29}) can be replaced by the weight matrix in the GCN, i.e., $\textbf{W}_{\text{conv2}}$, for reducing the number parameters when training HyperGCN. The remaining steps of Node2Vec-HyperGCN follow Node2Vec-SLP. 

Similarly to Node2Vec-GCN, Node2Vec-GraphSAGE, and Node2Vec-RGCN, other well-developed hypergraph-based neural networks such as \cite{yang2020hypergraph, bandyopadhyay2020line, arya2020hypersage, yi2020hypergraph, dong2020hnhn,zhang2021learnable} can also be applied for the step of embedding refinement.  Node2Vec-HyperGCN has achieved reasonable performances in hyperlink prediction, but it was initially developed for node classification/clustering, which focus more on using hyperlink relations to classify node labels.

\subsubsection{Hypergraph Learning over Families of Sets}
Hypergraph learning over families of sets (FamilySet) is a direct hyperlink prediction method which can learn provably expressive representations of hyperlinks with variable degrees that preserve local-isomorphism in the line graph of the hypergraph (Fig. \ref{fig:3} d) \cite{srinivasan2021learning}. FamilySet uses a message passing framework on the incidence graph representation of the incomplete hypergraph, which synchronously updates the node and hyperlink embeddings. Let $\textbf{x}_i$ and $\textbf{y}_p$ represent the embeddings of node $v_i$ and hyperlink $e_p$, respectively. Then the updating rule is given by
\begin{equation}\label{eq:30}
    \begin{split}
        \hat{\textbf{y}}_p &= \sigma\Big(\textbf{W}_{\text{conv1}}\Big(\textbf{y}_p || f\big( \{\textbf{x}_i || t(\{\textbf{y}_{p^{'}}\})\}\big)\Big) \Big)\\
        &\text{ for } v_i\in e_p \text{ and }  e_{p^{'}}\ni v_i\\
        \hat{\textbf{x}}_i &= \sigma \Big(\textbf{W}_{\text{conv2}} \Big(\textbf{x}_i || g\big( \{\textbf{y}_p || s(\{\textbf{x}_{i^{'}}\})\}\big)\Big)\Big)\\ &\text{ for }
         v_i\in e_p \text{ and } v_{i^{'}}\in e_{p},
    \end{split}
\end{equation}
where ``$||$" denotes the vector concatenation operation, $f, g, t, s$ are injective set functions, and $\textbf{W}_{\text{conv1}}$ and $\textbf{W}_{\text{conv2}}$ are learnable parameters in the convolutional networks. One may update the node and hyperlink embeddings for a certain number of iterations. The final representation of hyperlink $e_p$ is then given by
\begin{equation}
    \hat{\hat{\textbf{y}}}_p =  \phi(\{\hat{\textbf{x}}_i\}) || \rho (\{\hat{\textbf{y}}_{p^{'}}\}) \text{ for } v_i\in e_p \text{ and }  e_{p^{'}}\ni v_i ,
\end{equation}
where $\phi$ and $\rho$ are injective set and multiset functions, respectively.  In the end, FamilySet utilizes maximum likelihood estimation to estimate a classifier using the final hyperlink emebddings. Utilization of line graphs successfully enhances the interactions between hyperlinks with its local environment, which enables FamilySet to outperform the previous indirect methods such as Node2Vec-GraphSAGE, Node2Vec-RGCN, and Node2Vec-HGCN. 

\subsubsection{Structural Representation Neural Network and Local Spectrum}
Structural representation neural network and local spectrum (SNALS) is a hyperlink prediction method that exploits bipartite GCN with structural features (Fig. \ref{fig:3} c) \cite{wan2021principled}. Given a hyperlink $e_p$, define its $q$-hop neighbor node set $\mathcal{V}_q$, hyperlink set $\mathcal{E}_q$, and affinity matrix $\textbf{X}_q$ as follows:
\begin{align}
\begin{split}
    \mathcal{V}_q &= \{v_j | \eta(v_i, v_j)\leq q \text{ for } v_j\in \mathcal{V} \text{ and } v_i\in e_p\},\\
    \mathcal{E}_q &= \{e_s | e_s\subseteq \mathcal{V}_q \text{ and } e_s\in \mathcal{E}\},\\
    (\textbf{X}_q)_{ij} &= \eta(v_i, v_j|v_i\in \mathcal{V}_q \text{ and } v_j\in e_p)\in\mathbb{R}^{|\mathcal{V}_q|\times c_p},  
    \end{split}
\end{align}
where $\eta(v_i, v_j)$ denotes the shortest path distance between nodes $v_i$ and $v_j$. SNALS first integrates the affinity matrix $\textbf{X}_q$ using a bipartite GCN, i.e., generating and refining embeddings of the nodes in $\mathcal{V}_q$ by 
\begin{equation}\label{eq:32}
\begin{split}
    \textbf{X}_{\mathcal{V}_q} &= \text{setNN}(\textbf{X}_q),\\
    \textbf{X}_{\mathcal{E}_q} &= \sigma(\textbf{C}_q^{-1}\textbf{H}_q^\top \textbf{X}_{\mathcal{V}_q}\textbf{W}_{\text{conv1}}),\\
    \hat{\textbf{X}}_{\mathcal{V}_q} &= \sigma(\textbf{H}_q\textbf{X}_{\mathcal{E}_q}\textbf{W}_{\text{conv2}}),
\end{split}
\end{equation}
where setNN represents the set neural network for standardizing $\textbf{X}_q$ into a feature matrix of a fixed dimension,  $\textbf{C}_q$ and $\textbf{H}_q$ are the cardinality matrix and the incidence matrix of the $q$-hop hypergraph, respectively, and  $\textbf{W}_{\text{conv1}}$ and $\textbf{W}_{\text{conv2}}$ are learnable parameters in the bipartite GCN. One may update the $q$-hop node and hyperlink embeddings for a certain number of iterations. Then the embedding of hyperlink $e_p$, denoted by $\textbf{y}_p$, can be obtained by aggregating all the embeddings of the nodes within the hyperlink through a pooling function. In order to keep the structure of the affinity of $\textbf{X}_q$ in the representation of hyperlink $e_p$, SNALS further utilizes the top two singular values of $\textbf{X}_q$ to reflect the low rank property of the affinity matrix, i.e., the topological structure. Finally, SNALS feeds the combined features (i.e., $\textbf{y}_p$ and top two singular values) into a one-layer neural network to produce a hyperlink score as (\ref{eq:25}). SNALS can capture the joint interactions of a hyperlink by its local environment and overcome the both node-level and hyperlink-level ambiguity issues, see details in \cite{wan2021principled}.

\subsubsection{Deep Hypernetwork Embedding}
Deep hypernetwork embedding (DHNE) is a deep learning-based model that realizes a nonlinear tuple-wise similarity function while preserving both local and global proximities in the formed embedding space \cite{tu2018structural}. Given an incomplete uniform hypergraph $\mathcal{H}$ with $n$ nodes, DHNE initializes the node embeddings of $\mathcal{H}$ through an auto-encoder. The encoder is a nonlinear mapping from the adjacency space $\textbf{A}=\textbf{H}\textbf{H}^\top-\textbf{D}\in\mathbb{R}^{n\times n}$ to a latent representation space \textbf{X}, and the decoder is a nonlinear mapping from the latent representation \textbf{X} space back to the original adjacency space $\tilde{\textbf{A}}$, i.e., 
\begin{align}
\begin{split}
        \textbf{X} =\sigma(\textbf{A}\textbf{W}_{\text{enc}} + \textbf{b}_{\text{enc}}),\\
        \tilde{\textbf{A}} = \sigma(\textbf{X}\textbf{W}_{\text{dec}} + \textbf{b}_{\text{dec}}),
\end{split}
\end{align}
where $\tilde{\textbf{A}}$ is used to compute a reconstruction loss, and $\textbf{W}_{\text{enc}}$ and $\textbf{W}_{\text{dec}}$ are learnable parameters in the auto-encoder. Suppose that the embedding of node $v_i$ is $\textbf{x}_i$ (i.e., the $i$th row of \textbf{X}), DHNE computes a hyperlink score of a hyperlink $e_p$ through a multi-layer perceptron (MLP), i.e., 
\begin{equation}
    S_p = \text{sigmoid}\Big(\textbf{W}_{\text{score}}\sigma(\textbf{W}_{\text{linear}} \sum_{v_i\in e_p} \textbf{x}_i + \textbf{b}_{\text{linear}}) + \textbf{b}_{\text{score}}\Big),
\end{equation}
where $\textbf{W}_{\text{linear}}$, $\textbf{W}_{\text{score}}$, $\textbf{b}_{\text{linear}}$, $\textbf{b}_{\text{score}}$ are learnable parameters in the MLP.

DHNE directly models the tuple-wise relationship using an MLP and achieves a better performance on multiple tasks as compared to Node2Vec-SLP. However, the structure of the MLP takes fixed-size inputs, making DHNE only applicable to uniform hypergraphs. Moreover, using an MLP to refine node embedding fails to capture node-level interactions within each hyperlink.

\subsubsection{Self Attention-based Graph Convolutional Network for Hypergraphs}
Self attention-based graph convolutional network for hypergraphs (HyperSAGCN) generalizes DHNE by exploiting a SAGCN in refining embeddings of the nodes within each hyperlink \cite{zhang2019hyper}. HyperSAGCN offers two options in generating initial node embeddings. The first option is using the auto-encoder-based approach proposed in DHNE, while the second option is using a hypergraph random walk-based approach (see details in \cite{zhang2019hyper}). Suppose that the embedding of node $v_i$ is $\textbf{x}_i$ (obtained from either the two options). Given a hyperlink $e_p$, HyperSAGCN incorporates two different ways (static and dynamic) to refine the embeddings of the nodes within $e_p$, i.e., 
\begin{align}
\begin{split}
    \textbf{s}_i &= \sigma(\textbf{W}_{\text{linear}} \textbf{x}_i) \text{ for } v_i\in e_p,\\
    \textbf{d}_i &= \sigma(\sum_{\substack{v_i,v_j\in e_p \\ i\neq j}}\alpha_{ij}\textbf{W}_{\text{conv}}\textbf{x}_j),
\end{split}
\end{align}
where $\alpha_{ij}$ are the attention coefficients defined by
\begin{equation*}
    \alpha_{ij} = \frac{\exp{\Big((\textbf{W}_i^\top\textbf{x}_i)^\top (\textbf{W}_j^\top\textbf{x}_j)\Big)}}{\sum_{k=1}^{c_p}\exp{\Big((\textbf{W}_i^\top\textbf{x}_i)^\top (\textbf{W}_k^\top\textbf{x}_k)\Big)}},
\end{equation*}
and $\textbf{W}_{\text{linear}}$ and $\textbf{W}_{\text{conv}}$ are learnable parameters in the static and dynamic neural networks, respectively. The embedding of hyperlink $e_p$ through a mean pooling is then given by
\begin{equation}
    \textbf{y}_p = \frac{1}{c_p} \sum_{v_i \in e_p} (\textbf{s}_i - \textbf{d}_i)^{\otimes2},
\end{equation}
where the subscript $``\otimes2"$ denotes the element-wise square power. The final hyperlink score is same as (\ref{eq:25}). HyperSAGCN is able to improve the performance of DHNE  while addressing the shortcomings  such as the inability to predict hyperlinks for non-uniform hypergraphs. Yet,  HyperSAGCN does not perform well on relatively sparse hypergraphs such as metabolic networks. 

\subsubsection{Neural Hyperlink Predictor}
Neural hyperlink predictor (NHP) is an improved version of HyperSAGCN, where it employs a new maximum minimum-based pooling function which can adaptively learn weights in a task-specific manner and include more prior knowledge about the nodes \cite{yadati2020nhp}. Similar to those indirect methods, NHP initializes node embedding by performing Node2Vec on the clique-expanded graph. Suppose that the embedding of node $v_i$ is $\textbf{x}_i$. Given a hyperlink $e_p$, NHP treats it as a fully connected graph and refines the embeddings of the nodes within $e_p$ by a GCN defined in (\ref{eq:26}) (while Node2Vec-GCN applies a GCN on the entire expanded graphs). Then NHP uses a maximum minimum-based pooling function to compute hyperlink embeddings, i.e., 
\begin{equation}\label{eq:38}
    y^{\text{maxmin}}_{pl} = \max_{v_i\in e_p} \{\hat{x}_{il}\} - \min_{v_i\in e_p} \{\hat{x}_{il}\},
\end{equation}
where $y^{\text{maxmin}}_{pl}$ denotes the $l$th entry of $\textbf{y}_p^{\text{maxmin}}$, and $\hat{x}_{il}$ denotes the $l$th entry of the refined embedding $\hat{\textbf{x}}_i$. The final scoring function is same as (\ref{eq:25}). The employment of the maximum minimum-based pooling function enables NHP to outperform the previous methods such as Node2Vec-SLP, Node2Vec-HGCN, Node2Vec-HyperGCN, and HyperSAGCN. However, NHP does not perform well on relatively dense hypergraphs such as email communication networks (due to utilization of Node2Vec), contrary to HyperSAGCN.

\subsubsection{Chebyshev Spectral Hyperlink Predictor}
Chebyshev spectral hyperlink predictor (CHESHIRE) is a recent hyperlink prediction method built upon HyperSAGCN and NHP \cite{chen2022teasing}. CHESHIRE has been successfully applied to predict missing reactions in genome-scale metabolic networks, which significantly improves the phenotype prediction of metabolic models. Different from NHP and HyperSAGCN, CHESHIRE initializes node embedding by simply passing the incidence matrix $\textbf{H}$ through a one-layer neural network, i.e., 
\begin{equation}
    \textbf{x}_i = \sigma(\textbf{W}_{\text{enc}}\textbf{h}_i + \textbf{b}_{\text{enc}}),
\end{equation}
where $\textbf{h}_i$ is the $i$th row of \textbf{H}, and $\textbf{W}_{\text{enc}}$ and $\textbf{b}_{\text{enc}}$ are learnable parameters in the encoder. Node2Vec on the clique expansion could lose higher-order structural information and require a great amount of computation resources, while the adjacency matrix of a hypergraph is in fact equivalent to the adjacency matrix of its clique expansion. Initialization with the incidence matrix is able to preserve multi-dimensional relationships while keeping low memory costs. 

Suppose that the embedding of node $v_i$ is $\textbf{x}_i$. Given a hyperlink $e_p$, CHESHIRE treats it as a fully connected graph (as HyperSAGCN and NHP) and refines the embeddings of the nodes within $e_p$ with a Chebyshev spectral GCN, i.e., 
\begin{equation}
    \hat{\textbf{x}}_i = \sigma\Big(\sum_{k=1}^K \textbf{W}_{\text{conv}}^{(k)} \textbf{z}^{(k)}_i \Big) \text{ for } v_i\in e_p,
\end{equation}
where $K$ is the Chebyshev filter size, $\textbf{z}^{(k)}_i$ are computed recursively by
\begin{equation*}
    \textbf{z}_i^{(1)} = \textbf{x}_i, \text{ } \textbf{z}_i^{(2)} = \tilde{\textbf{L}}_\text{c}\textbf{x}_i, \text{ and } \textbf{z}^{(k)}_i = 2\tilde{\textbf{L}}_\text{c}\textbf{z}_i^{(k-1)}-\textbf{z}_i^{(k-2)},
\end{equation*}
and $\textbf{W}_{\text{conv}}^{(k)}$ are learnable parameters in the Chebyshev spectral GCN. The matrix $\tilde{\textbf{L}}_\text{c}$ is the scaled normalized Laplacian matrix of the fully connected graph defined by
\begin{equation*}
    \tilde{\textbf{L}}_\text{c} = \frac{2}{\lambda_{\max}} \textbf{L}_\text{c} - \textbf{I} = \frac{2}{\lambda_{\max}}(\textbf{I} - \textbf{D}_\text{c}^{-\frac{1}{2}}\textbf{A}_\text{c}\textbf{D}_\text{c}^{-\frac{1}{2}}) - \textbf{I},
\end{equation*}
where $\textbf{L}_\text{c}$ is the symmetric normalized Laplacian matrix of the graph with the largest eigenvalue $\lambda_{\max}$, and $\textbf{D}_\text{c}$ and $\textbf{A}_\text{c}$ are the degree matrix and the adjacency matrix of the graph, respectively. The Chebyshev spectral GCN exploits the Chebyshev polynomial expansion and spectral graph theory to learn the localized spectral filters, which can extract local and composite features on graphs that encode complex geometric structures \cite{defferrard2016convolutional}.

\begin{figure*}[t]
    \centering
    \includegraphics[scale=1.4]{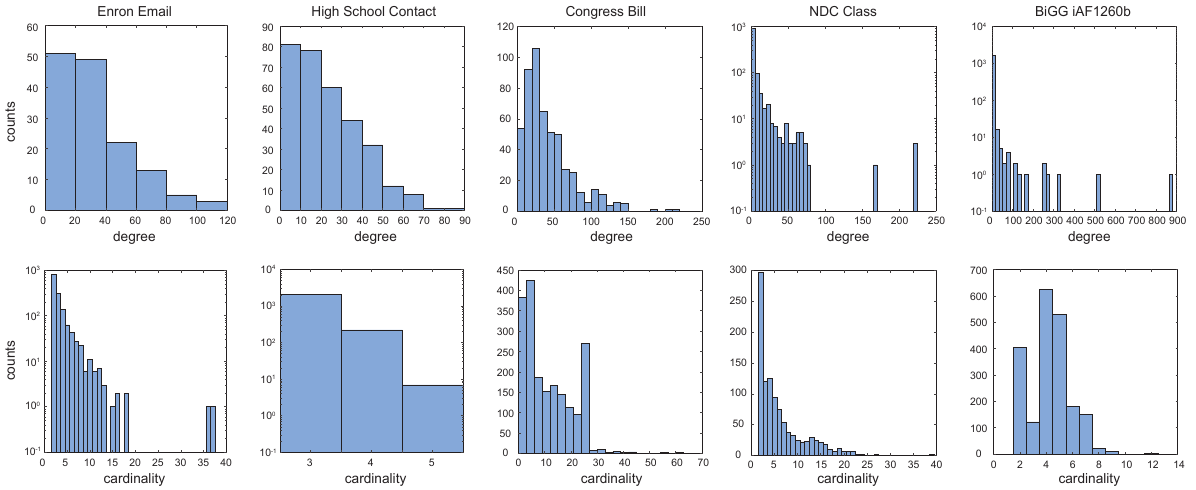}
    \caption{Node degree distribution and hyperlink cardinality distribution of the five hypergraphs.}
    \label{fig:4}
\end{figure*}

Subsequently, CEHSHIRE employs a Frobenius norm-based (also known as the 2-norm) pooling function to generate hyperlink embeddings, i.e.,
\begin{equation}
    y^{(\text{norm})}_{pl} = \Big(\frac{1}{c_p}\sum_{v_i \in e_p} \hat{x}_{il}^2\Big)^{\frac{1}{2}},
\end{equation}
where $y^{\text{norm}}_{pl}$ denotes the $l$th entry of $\textbf{y}_p^{\text{norm}}$, and $\hat{x}_{il}$ denotes the $l$th entry of $\hat{\textbf{x}}_i$. The Frobenius norm-based function is  efficient at separating boundaries of hyperlink embedding space \cite{gulcehre2014learned}. In order to achieve a better performance, CHESHIRE also incorporates the maximum minimum-based pooling function as defined in (\ref{eq:38}). Thus, the final score of hyperlink $e_p$ is given by
\begin{equation}
    S_p = \text{sigmoid}\Big(\textbf{W}_{\text{score}} (\textbf{y}_p^{(\text{norm})} || \textbf{y}_p^{(\text{maxmin})}) + \textbf{b}_{\text{score}}\Big),
\end{equation}
where ``$||$'' denotes the vector concatenation operation, and 
$\textbf{W}_{\text{score}}$ and $\textbf{b}_{\text{score}}$ are learnable parameters in the neural network. CHESHIRE  has effectively addressed the limitations of HyperSAGCN and NHP and achieved an outstanding performance on various types of hypergraph data.

\section{Experiments}\label{sec:4}
We selected representative methods from each category to perform a benchmark study on multiple hypergraph applications. Since those indirect methods were proposed as baseline methods in the work of hyperlink prediction, we only considered direct methods here. The selected methods include HPRA (similarity-based), HPLSF (probability-based), C3MM (matrix optimization-based), HyperSAGCN (deep learning-based), NHP (deep learning-based), and CHESHIRE (deep learning-based). We did not compare the matrix optimization-based approach HPLS and the other two deep learning-based approaches FamilySet and SNALS since their codes are not publicly available. We used the second version of HPLSF described in \cite{zhang2018beyond}. We tried to use the same hyperparameters as set by default in the original codes. For HyperSAGCN, NHP, and CHESHIRE, we set their hidden dimensions to be consistent and utilized the same loss function defined in \cite{liu2017computational}. All the experiments presented were performed on a Macintosh machine with 32 GB RAM and an Apple M1 Pro chip in Python and MATLAB R2022a.

\begin{table}[t]
\centering
\caption{A summary of  Enron email, high school contact, congress bill, NDC class, and BiGG metabolic networks. Note that the statistics may differ from the original datasets since we pre-processed the networks (by deleting duplicated hyperlinks and hyperlinks with cardinality one).}
\begin{tabularx}{8cm}{STST}
\toprule
\centering\arraybackslash Dataset  & \centering\arraybackslash \# Nodes  & \centering\arraybackslash \# Hyperlinks &  \centering\arraybackslash Reference \\
\midrule
\centering\arraybackslash Enron Email & \centering\arraybackslash 143 & \centering\arraybackslash 1,459 & \centering\arraybackslash \cite{benson2018simplicial}\\
\centering\arraybackslash High School & \centering\arraybackslash 317 & \centering\arraybackslash 2,320 & \centering\arraybackslash \cite{benson2018simplicial}\\
\centering\arraybackslash Congress Bill & \centering\arraybackslash 531 & \centering\arraybackslash 1,973 & \centering\arraybackslash \cite{benson2018simplicial}\\
\centering\arraybackslash NDC Class & \centering\arraybackslash 1,149 & \centering\arraybackslash 1,049 & \centering\arraybackslash \cite{benson2018simplicial}\\
\centering\arraybackslash BiGG iAF1260b & \centering\arraybackslash 1,668 & \centering\arraybackslash 2,046 & \centering\arraybackslash \cite{king2016BiGG}\\
\bottomrule
\end{tabularx}
\label{tab:3}
\end{table}

\begin{table*}[t]
\centering
\caption{Test results for the Enron email, high school contact, congress bill, NDC class, and BiGG metabolic networks using the metrics AUROC and F1 scores. We chose $\alpha=0.5$ and $\beta=1$ in negative sampling. Each value is the mean over 10 trials.}
\vspace{0.3cm}
\begin{tabularx}{16cm}{c *{10}{Y}}
\toprule
Dataset
 & \multicolumn{2}{c}{Enron Email} 
 & \multicolumn{2}{c}{High School }
 & \multicolumn{2}{c}{Congress Bill}
 & \multicolumn{2}{c}{NDC Class}
 & \multicolumn{2}{c}{BiGG iAF1260b}\\
\cmidrule(lr){2-3} \cmidrule(l){4-5} \cmidrule(l){6-7} \cmidrule(l){8-9} \cmidrule(l){10-11}
  & AUROC  & F1   & AUROC  & F1  & AUROC  & F1   & AUROC  & F1  & AUROC  & F1  \\
\midrule
HPRA  & 0.8486   & 0.7017  & 0.9694 & 0.8456  &0.7957  & 0.6596  & 0.7165 & 0.5110  & 0.7323 & 0.5962\\
HPLSF  & 0.7657   & 0.7020  & 0.9223 & 0.8433  & 0.8073 & 0.7500  & 0.7623 & 0.7152  & 0.6418 & 0.5760\\
C3MM  &  0.7207  & 0.6885  & 0.8839 & 0.7984  & 0.7178 & 0.6895  &0.6800  & 0.6501  &0.7227 & 0.6770\\
HyperSAGCN  & \textbf{0.8799}   & \textbf{0.8291}  & 0.9637 & 0.9291  & 0.8421 & 0.7909  & 0.8190 & 0.6720  & 0.6606 & 0.5407\\
NHP  & 0.6578   & 0.6202  & 0.7755 & 0.7121  & 0.8086 & 0.7253  & 0.8321  & 0.7745  &0.8134  & 0.7311\\
CHESHIRE  & 0.8608   & 0.7914  & \textbf{0.9792} & \textbf{0.9360}  & \textbf{0.9271} & \textbf{0.8606}  & \textbf{0.8869}  &  \textbf{0.8149} & \textbf{0.8418} & \textbf{0.7697}\\
\bottomrule
\end{tabularx}
\label{tab:4}
\end{table*}

\begin{table*}[t]
\centering
\caption{Test results for the Enron email, high school contact, congress bill, NDC class, and BiGG metabolic networks using the metrics AUROC and F1 scores. We chose $\alpha=0.8$ and $\beta=3$ in negative sampling. Each value is the mean over 10 trials.}
\vspace{0.3cm}
\begin{tabularx}{16cm}{c *{10}{Y}}
\toprule
Dataset
 & \multicolumn{2}{c}{Enron Email} 
 & \multicolumn{2}{c}{High School}
 & \multicolumn{2}{c}{Congress Bill}
 & \multicolumn{2}{c}{NDC Class}
 & \multicolumn{2}{c}{BiGG iAF1260b}\\
\cmidrule(lr){2-3} \cmidrule(l){4-5} \cmidrule(l){6-7} \cmidrule(l){8-9} \cmidrule(l){10-11}
  & AUROC  & F1   & AUROC  & F1  & AUROC  & F1   & AUROC  & F1  & AUROC  & F1  \\
\midrule
HPRA  & 0.7405   & 0.4872  & 0.8173 & 0.5831  &0.6662  & 0.4235  & 0.6140 & 0.3457  & 0.6066 & 0.3839\\
HPLSF  & 0.7696   & 0.5642  & \textbf{0.9193} & \textbf{0.7422}  & \textbf{0.8397} & \textbf{0.6352}  & 0.7633 & 0.5693  & 0.6178 & 0.3967\\
C3MM  &  0.6270  & 0.4351  & 0.8163 & 0.5949  & 0.5987 & 0.3864  &0.6666  & 0.4496  &0.6724 & 0.4353\\
HyperSAGCN  & \textbf{0.8588}   & \textbf{0.6673}  & 0.9081 & 0.7338  & 0.7879 & 0.5417  & 0.7902 & 0.5548  & 0.5257 & 0.3137\\
NHP  & 0.5808   & 0.4087  & 0.6319 & 0.4316  & 0.6391 & 0.4240  & 0.7375  & 0.5256  &0.6465  & 0.4382\\
CHESHIRE  & 0.8271   & 0.6253  & 0.9042 & 0.7419  & 0.8280 & 0.5881  & \textbf{0.8181}  &  \textbf{0.6092} & \textbf{0.6816} & \textbf{0.4711}\\
\bottomrule
\end{tabularx}
\label{tab:5}
\end{table*}
\subsection{Benchmark Datasets}
We used the following five  datasets to have direct comparisons among HPRA, HPLSF, C3MM, HyperSAGCN, NHP, and CHESHIRE:
\begin{itemize}
    \item Enron email network, where each employee is a node and each email represents a hyperlink connecting the sender and the recipients in the email. 
    \item High school contact network, where each student/teacher is a node and each face-to-face contact represents a hyperlink connecting the people involved in the contact. We only considered contacts that involve more than two persons in this network.
    \item Congress bill network, where each US Congressperson is a node and each legislative bill put forth in both the House of Representatives and the Senate represents a hyperlink connecting the sponsors and co-sponsors of the bill. We only considered bills put forth within a range of time. 
    \item NDC class network, where each class label is a node and each drug represents a hyperlink connecting the class labels of the drug.
    \item BiGG metabolic network (E. coli Model iAF1260b), where each metabolite is a node and each chemical reaction represents a hyperlink connecting the reactant and product metabolites in the reaction.
\end{itemize}
For all the networks, we did not consider duplicated hyperlinks or hyperlinks with cardinality one. Details of the datasets, including the number of nodes and the number of hyperlinks, are shown in Table \ref{tab:3}. Node degree and hyperlink cardinality distributions of each dataset are shown in Fig. \ref{fig:4}. The BiGG metabolic network (iAF1260b) can be downloaded from the BiGG database \cite{king2016BiGG}, and the remaining datasets can be downloaded from \cite{benson2018simplicial}.

\subsection{Negative Sampling}
Most of the selected hyperlink prediction methods including HPLSF, HyperSAGCN, NHP, and CHESHIRE require negative sampling, i.e., creating fake hyperlinks that do not exist, during training to balance specificity and sensitivity of the trained model. We here generalize the sampling strategy proposed in NHP. Suppose that we have a hypergraph $\mathcal{H}$. For each (positive) hyperlink $e\in \mathcal{E}$, we generate a corresponding negative hyperlink $f$, where $\alpha\times 100 \%$  of the nodes in $f$ are from $e$  and the remaining are from $\mathcal{V}\backslash \{e\}$. Denote the negative hyperlink set as $\mathcal{F}$. The number $\alpha$ controls the genuineness of the negative hyperlinks, i.e., higher values of $\alpha$ indicate that the negative hyperlinks are more close to the true. Additionally, we define $\beta$ to be the number of times of negative sampling which controls the ratio between positive and negative hyperlinks. Note that the remaining two methods - HPRA and C3MM do not require negative sampling, but we intentionally introduced negative hyperlinks to the testing/candidate set in order to have a fair comparison. More negative sampling strategies used for hyperlink prediction can be found in \cite{patil2020negative}.

\subsection{Evaluation Metrics}
We evaluated the performance based on the area under the receiver operating characteristic curve (AUROC) and F1 score. AUROC is one of the most popular evaluation metrics for checking any classification model’s performance. AUROC can tell how much the model is capable of distinguishing between classes. The higher the AUROC, the better the model is at predicting 0 classes as 0 and 1 classes as 1. On the other hand, F1 score combines recall and precision of a classifier into a single metric by taking their harmonic mean. Recall measures the model's ability to detect positive samples, while precision reflects how reliable the model is in classifying samples as positive. Both the metrics can represent the overall capability of each hyperlink prediction method in recovering missing hyperlinks.

\subsection{Results}
We first considered a negative sampling strategy with $\alpha=0.5$ and $\beta=1$ as used in NHP.  For each network, we  randomly split the hyperlink set including positive and negative hyperlinks into training and testing sets by a ratio of 3:2 over 10 trials (no negative sample was introduced in the training set for HPRA and C3MM). We trained the six learning models on the training set and tested on the testing set.  The results are shown in Table \ref{tab:4}, where each value is the mean over 10 trials, and we picked the mean of all the testing hyperlink scores as the cutoff threshold for computing the F1 score. We found that CHESHIRE achieves an overall preeminent and stable performance on the five networks compared to the other methods, despite being slightly lower than the performance of HyperSAGCN on the Enron email network. In addition, HPRA and HyperSAGCN perform outstandingly on the relatively dense networks (e.g., the Enron email and high school contact networks), competitive with CHESHIRE. However, they have poor performances on the relatively sparse networks (e.g., the NDC class and BiGG metabolic networks). Interestingly, the performance of  NHP is completely contrary to these of HPRA and HyperSAGCN. NHP initiates node emebeddings of a hypergraph by using Node2Vec on its expanded graph. Thus, when the hypergraph becomes dense, the graph-based node embeddings may not be able to accurately capture the structural features of the hypergraph. Lastly, both HPLSF and C3MM have mediocre performances on all the networks. 

We repeated the same experiment using another negative sampling parameters with $\alpha=0.8$ and $\beta=3$. The results are shown in Table \ref{tab:5}, where the six hyperlink prediction methods behave similarly as the previous, despite considerable decreases due to the introduction of more genuine negative hyperlinks. Notably, the performance of HPLSF is least affected by the change of the negative sampling parameters, where the AUROC scores are almost consistent under the two settings. It even achieves the best performance on the high school contact and congress bill networks. On the other hand, the presence of more realistic negative samples severely undermines the performance of HPRA. Last but not least, the deep learning-based model CHESHIRE still accomplishes a superb and stable performance over all the networks.  

\begin{table}[t]
\centering
\caption{Total computational time of the selected methods for running the five datasets sequentially on a Macintosh machine with 32 GB RAM and an Apple M1 Pro chip in Python and MATLAB R2022a.}
\begin{tabularx}{6cm}{c *{2}{Y}}
\toprule
Method  & Time (s)    \\ 
\midrule
HPRA & 0.7642 \\ 
HPLSF & 8.0214  \\ 
C3MM & 462.8452 \\ 
HyperSAGCN & 144.4734 \\
NHP &  149.4622\\
CHESHIRE & 178.7379\\
\bottomrule
\end{tabularx}
\label{tab:6}
\end{table}

\section{Discussion}\label{sec:5}
The experiments reported here highlight that the six selected direct hyperlink prediction methods are able to effectively recover artificially removed hyperlinks. We also computed the total computational time of each method for running the five datasets at once based on the first negative sample rule (Table \ref{tab:6}). We have the following interesting observations according to their experiment performances: 
\begin{itemize}
    \item HPRA is the most computationally efficient method among the six selected methods. It only took less a second to finish all the testing networks. More importantly, HPRA performs well on relatively dense hypergraphs (e.g., the Enron email and high school contact networks). Hence, HPRA will be extremely useful when handling hyperlink prediction on large dense hypergraphs.
    \item HPLSF is also a computationally efficient method. It only costs around 8 seconds to finish all the testing networks. Surprisingly, HPLSF exhibits a strong performance when dealing with more  genuine negative hyperlinks, particularly on those dense hypergraphs. 
    \item C3MM is the most time-consuming method among the six selected methods. It took roughly 8 minutes to finish all the testing networks. The main reason of such slow speed is that C3MM requires the testing/candidate hyperlink set to be present during training. In the meantime, C3MM cannot achieve a competitive performance compared to other methods even if it takes more time. 
    \item The computational time of CHESHIRE is about 3 minutes for completing all the testing networks. CHESHIRE achieves the most robust performance compared to the other selected methods, i.e., the AUROC and F1 scores are consistently outstanding from dense hypergraphs to sparse hypergraphs (even though they may not be the top one). Moreover, the deep learning-based approaches (including HyperSAGCN, NHP, and CHESHIRE) still exhibit considerable advantages over other methods for hyperlink prediction. 
\end{itemize}

However, we were unable to compare the other two deep learning-based approaches - FamilySet and SNALS due to the lack of code availability. It will be useful to include the two methods in future comparisons. Furthermore, we believe that deep learning-based approaches are more promising in predicting missing hyperlinks on hypergraphs. Particularly, most hypergraph-based neural networks aim to decompose a hypergraph into a graph (such as clique expansion, star expansion, and line expansion) while keeping its higher-order topological attributes. Thus, how to construct expanded graphs from hypergraphs and how to select appropriate graph-based neural networks running on the expanded graphs become important in designing a new learning architecture for hyperlink prediction. Other than the four mentioned categories, the idea of hypergraphon, a generalization of graphons, might be potentially powerful for hyperlink prediction \cite{borgs2017graphons, zhao2015hypergraph}. 

In this paper, we systematically and comprehensively reviewed recent progresses on hyperlink prediction. In particular, we proposed a new taxonomy to classify existing hyperlink prediction methods into four categories which include similarity-based, probability-based, matrix optimization-based, and deep learning-based methods.  To study the effectiveness of different types of the methods, we further conducted a benchmark study on various hypergraph applications, including email communication, school contact, congress bill, drug class, and metabolic networks, using representative methods from each category. As mentioned above, deep learning-based methods still prevail over other conventional methods.  It will be worthwhile to develop a novel and higher-order-preserved hypergraph decomposition with appropriate graph-based neural networks for facilitating the performance of hyperlink prediction.


\ifCLASSOPTIONcaptionsoff
  \newpage
\fi




\bibliographystyle{IEEEtran}

\bibliography{citation}

\end{document}